\documentclass[10pt,twocolumn,letterpaper]{article}

\usepackage[pagenumbers]{iccv} 

\definecolor{iccvblue}{rgb}{0.21,0.49,0.74}
\usepackage[pagebackref,breaklinks,colorlinks,allcolors=iccvblue]{hyperref}
\usepackage{rotating}

\def\paperID{12703} 
\def\confName{ICCV}
\def\confYear{2025}

\title{CAM-Seg: A Continuous-valued Embedding Approach for Semantic Image Generation}

\author{Masud Ahmed\textsuperscript{$\star \P$}, Zahid Hasan\textsuperscript{$\star \P$}, Syed Arefinul Haque\textsuperscript{$\S$}, Abu Zaher Md Faridee \textsuperscript{$\star \dagger$}, \\ Sanjay Purushotham\textsuperscript{$\star$}, Suya You\textsuperscript{$\ddagger$}, Nirmalya Roy\textsuperscript{$\star \P$}\\
\textsuperscript{$\star$}\textit{University of Maryland Baltimore County}, USA\\
\textsuperscript{$\S$}\textit{Northeastern University}, USA\\
\textsuperscript{$\dagger$}\textit{Amazon Inc.}, USA \\
\textsuperscript{$\ddagger$}\textit{DEVCOM Army Research Laboratory}, USA \\
\textsuperscript{$\P$}\{mahmed10, zhasan3, nroy\}@umbc.edu
}

\begin{document}
\maketitle

\begin{abstract}


Traditional transformer-based semantic segmentation relies on quantized embeddings. However, our analysis reveals that autoencoder accuracy on segmentation mask using quantized embeddings (e.g. VQ-VAE) is 8\% lower than continuous-valued embeddings  (e.g. KL-VAE). Motivated by this, we propose a continuous-valued embedding framework for semantic segmentation. By reformulating semantic mask generation as a continuous image-to-embedding diffusion process, our approach eliminates the need for discrete latent representations while preserving fine-grained spatial and semantic details. Our key contribution includes a diffusion-guided autoregressive transformer that learns a continuous semantic embedding space by modeling long-range dependencies in image features. Our framework contains a unified architecture combining a VAE encoder for continuous feature extraction, a diffusion-guided transformer for conditioned embedding generation, and a VAE decoder for semantic mask reconstruction. Our setting facilitates zero-shot domain adaptation capabilities enabled by the continuity of the embedding space. Experiments across diverse datasets (e.g., Cityscapes and domain-shifted variants) demonstrate state-of-the-art robustness to distribution shifts, including adverse weather (e.g., fog, snow) and viewpoint variations. Our model also exhibits strong noise resilience, achieving robust performance ($\approx$ 95\% AP compared to baseline) under gaussian noise, moderate motion blur, and moderate brightness/contrast variations, while experiencing only a moderate impact ($\approx$ 90\% AP compared to baseline) from 50\% salt and pepper noise, saturation and hue shifts. Code available: \url{https://github.com/mahmed10/CAMSS.git}

\end{abstract}

\section{Introduction}
Traditional semantic segmentation is formulated as a pixel-wise classification task, typically relying on convolutional neural networks (CNNs) with fully connected layers~\cite{lecun2015deep}. Recent advancements~\cite{cheng2021per} propose an alternative formulation—predicting an entire segmentation map rather than classifying individual pixels. Alternatively, this approach can be defined as a conditional mask generation problem, where the model learns to produce a segmentation map conditioned on the input RGB image. Instead of directly classifying pixels, this formulation transforms the task into an image-to-image generation problem, where each pixel in the generated output represents a class through color palettes. Inspired by recent advancements in generative modeling~\cite{chen2023generative}, we introduce a transformer-based generative semantic segmentation framework that models segmentation as an autoregressive sequence generation task.

\begin{figure}[!t]
\centering
\includegraphics[width=\linewidth]{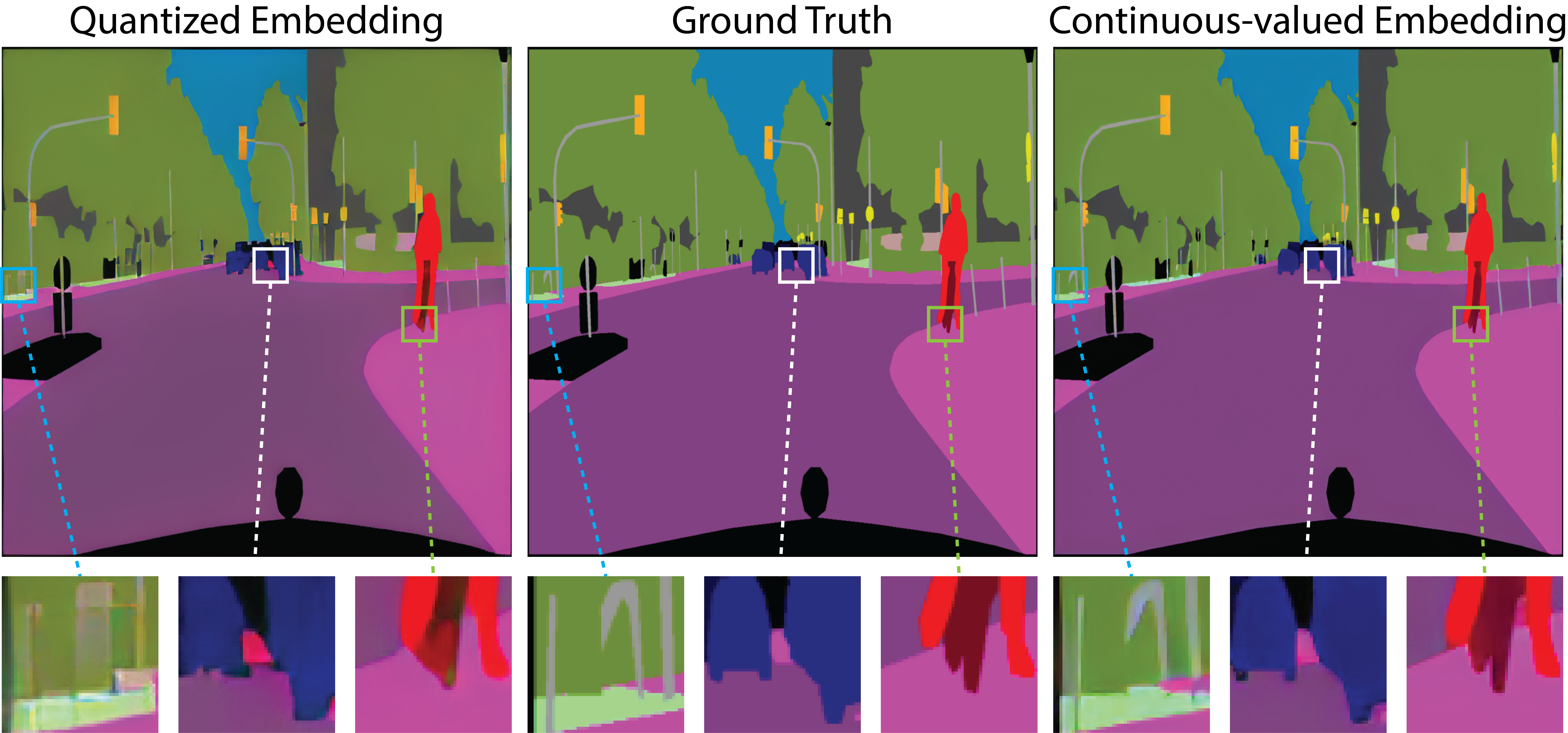}
\caption{Comparison of fine-grained details captured by quantized and continuous-valued embedding autoencoders. The blue box highlights discrepancies in pole reconstruction, the white box indicates inconsistencies in sidewalk representation, and the green box shows differences in the bicycle tire.}
\label{fig:intro}
\end{figure}

In traditional transformer-based generative models for computer vision, a widely adopted approach is discretizing image data~\cite{chen2020generative, ramesh2021zero, esser2021taming}. This is commonly achieved by training a quantized tokenizer, which converts images into a finite vocabulary of tokens using vector quantization (VQ)~\cite{van2017neural, razavi2019generating}. While this tokenization strategy simplifies modeling and sampling, it introduces information loss, limiting the model's ability to capture fine-grained details. As demonstrated in Fig.~\ref{fig:intro}, continuous-valued embeddings preserve richer feature representations and mitigate quantization-induced information loss, leading to finer-grained segmentation outputs. However, quantized embeddings remain more prevalent due to their compatibility with established probability distribution modeling techniques. In categorical distributions, two properties make quantized embeddings straightforward to use: (i) a well-defined loss function (e.g., cross-entropy~\cite{shannon1948mathematical}) for optimization, and (ii) a simplified sampling mechanism (e.g., Gumbel-max~\cite{gumbel1954statistical}). These properties make quantized embeddings easier to train and sample from, whereas continuous-valued embeddings require alternative loss formulations and sampling mechanisms for effective integration into generative models.

To address this issue, Li et al.~\cite{li2025autoregressive} propose training an autoregressive (AR) model on continuous-valued embeddings using a diffusion model, removing the need for quantization by employing a diffusion-based loss function with a strong result in synthetic RGB image generation. However, this method is limited to synthetic RGB image generation and was not originally designed for semantic segmentation tasks. Building upon this framework, we propose CAM-Seg, a Continuous Autoregressive Model (CAM) for semantic segmentation, incorporating bidirectional attention inspired by Masked Autoencoders (MAE)~\cite{he2022masked} to enhance feature learning and segmentation accuracy. The key contributions of our work are as follows.

\begin{itemize}
    \item \textbf{Novel Architecture:} We introduce a novel RGB image-conditioned semantic image generation approach that leverages a diffusion loss objective to train the transformer and diffusion model to generate continuous-valued semantic image embeddings from continuous-valued RGB image embeddings. It efficiently leverages pre-trained image autoencoders, requiring training only on the transformer and diffusion model, and enables multiple configurations. 
    
    \item \textbf{Noise Robustness \& Zero-Shot Adaptation:} Our model demonstrates robustness against various image degradations, sensor noise, motion artifacts, and lighting variations to maintain high-quality semantic embedding generation. It also excels in zero-shot domain adaptation, generalizing effectively across diverse environments, including adverse weather (fog, rain, snow), different image resolutions, and varying camera viewpoints. 
    
    \item \textbf{Comprehensive Experiments \& Ablation Studies:} We conduct extensive evaluations on four publicly available diverse datasets, analyzing model performance through ablation studies on input image size, semantic image color mapping, and data distribution shifts. These experiments provide valuable insights for optimizing model robustness and segmentation accuracy across different conditions.    
\end{itemize}

\section{Related Works}
Building on the success of transformers in NLP~\cite{vaswani2017attention}, vision transformers (ViT)~\cite{dosovitskiy2020image} utilize quantized embedding and patch-based attention to capture global context, making them more effective than traditional convolution-based methods for segmentation tasks. This lead to multiple generative models for semantic segmentation, such as Taming Transformers~\cite{esser2021taming}, Segmenter~\cite{strudel2021segmenter}, Mask2Former~\cite{cheng2022masked}, SETR~\cite{zheng2021rethinking}, SegFormer~\cite{xie2021segformer}, ClustSeg~\cite{liang2023clustseg}, OneFormer~\cite{jain2023oneformer}, and LOCATER~\cite{liang2023local}. Universal segmentation models such as SAM~\cite{kirillov2023segment} and SEEM~\cite{zou2024segment} have emerged as state of art, while few-shot approaches such as SegGPT~\cite{wang2023seggpt} leverage foundation models like CLIP~\cite{radford2021learning} and self-supervised learning methods such as DINO~\cite{caron2021emerging} and DINOv2~\cite{oquab2023dinov2}.

These generative models differ in their approach in representing images, using either a quantized embedding or a diffusion process. Techniques such as Stable Diffusion~\cite{rombach2022high}, autoregressive transformers without vector quantization~\cite{li2025autoregressive}, Maskbit~\cite{weber2024maskbit}, Llama for
Scalable Image Generation~\cite{sun2024autoregressive}, DART~\cite{gu2024dart} emerge as alternatives to quantized embedding. Generative Infinite-Vocabulary Transformers (GIVT)~\cite{tschannen2024givt} propose scalable generation strategies, while recent studies\cite{pasini2024continuous} highlight the advantages of continuous-valued embedding spaces for generative modeling. Continuous Autoregressive Models (CAMs)~\cite{pasini2024continuous} utilize diffusion to mitigate error accumulation in non-quantized autoregressive models~\cite{sun2024autoregressive}. Despite these advancements, none of the existing works have explored continuous-valued embedding-based autoregressive transformers for semantic segmentation. This paper aims to bridge this gap by introducing a novel approach that leverages continuous-valued representation learning for enhanced generalization and adaptability in semantic segmentation.

\section{Methodology} 
\begin{figure*}[!t]
    \centering
    \begin{minipage}{0.48\textwidth}
        \centering
        \includegraphics[width=\linewidth]{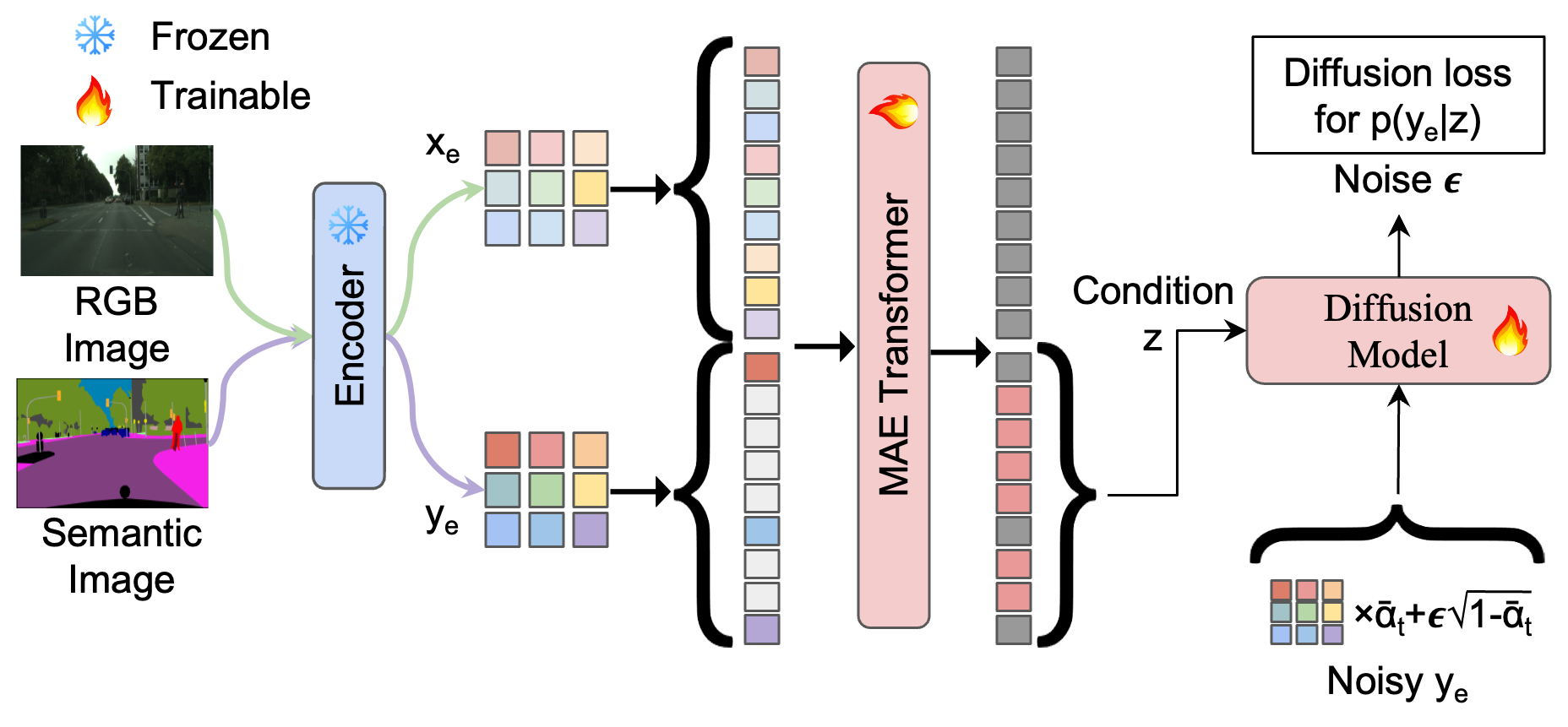}
        \subcaption{Training phase}
    \end{minipage}
    \begin{minipage}{0.01\textwidth}
        \centering
        \vrule width 1pt height 120pt 
    \end{minipage}
    \begin{minipage}{0.48\textwidth}
        \centering
        \includegraphics[width=\linewidth]{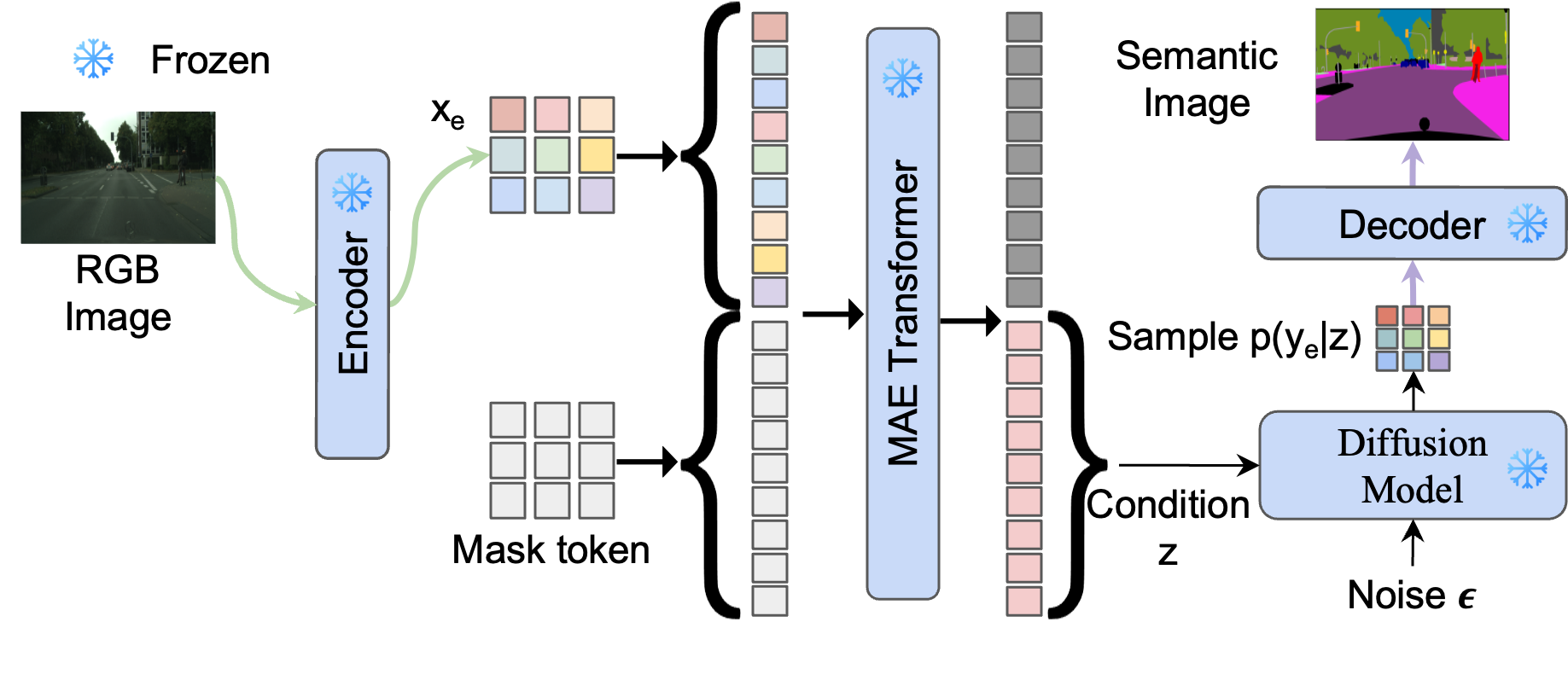}
        \subcaption{Inference phase}
    \end{minipage}
    \caption{Overview of CAM-Seg for semantic segmentation using a continuous-valued embedding-based autoregressive transformer. Our approach employs an autoencoder to extract continuous-valued embedding, which are processed by a transformer to generate conditioning variables for the diffusion model. The diffusion model then utilizes these conditions to compute loss (during training phase) or sample semantic embeddings (during inference phase), which are subsequently decoded back to the segmentation map space.}
    \label{fig:overall}
\end{figure*}

The overall pipeline is illustrated in Fig.~\ref{fig:overall}.

\textbf{Preliminary}
We represent the input RGB image as \(\mathbf{x}_i \in \mathbb{R}^{H \times W \times 3}\) and
ground truth segmentation map \(y_m \in \{0,1\}^{H\times W \times K}\), where each binary image corresponding to $K$ classes. Traditional approaches formulate this task as a pixel-wise classification problem, aiming to predict the binary class labels directly from the input image. However, to enable an image-space formulation, we adopt an inpainting-based approach~\cite{bar2022visual} and utilize the Cityscapes color palette~\cite{bar2022visual} to transform the ground truth map into a semantic colorized image \(\mathbf{y}_i \in \mathbb{Z}^{H \times W \times 3}\), where distinct color shades represent different semantic classes. The main objective of our approach is to predict \(y_i\) from \(x_i\) using a transformer-based model, which is subsequently mapped back to segmentation map \(y_m\). This setting aligns with the fixed $\beta$ matrix for Maskige estimation \cite{chen2023generative}.

\subsection{Continuous-valued Embedding Generation}
We employ an autoencoder to obtain $Z$ dimensional continuous-valued image embeddings ($\mathbf{x}_e$ and $\mathbf{y}_e$) from RGB images ($\mathbf{x}_i$) and its corresponding semantic image $\mathbf{y}_i$ using VAE feature encoder (\(\mathcal{E}: \mathbb{R}^{H\times W\times 3} \rightarrow \mathbb{R}^{\frac{H}{f}\times \frac{W}{f}\times Z} \)). We reshape the embedding to $L\times Z$ dimension with sequence length $L = \frac{H}{f}* \frac{W}{f}$, where $*$, $f$ denotes value multiplication and compression ration. Subsequently, the decoder \(\mathcal{D}\) reconstructs the original image from the embedding space. 

\textbf{Implementation Details:}
We adopt a KL-regularized autoencoder architecture similar to the Latent Diffusion Model (LDM)~\cite{rombach2022high}. This autoencoder model is trained using perceptual loss~\cite{zhang2018unreasonable}, and patch-based~\cite{isola2017image} adversarial objective~\cite{esser2021taming}. Following standard configurations, we set the downsampling factor to \(16\), resulting in a context length of \( L = 2304 \). The embedding dimension is set to \( Z = 16 \), ensuring a compact yet expressive latent space. To ensure generalization beyond domain-specific datasets, we initialize our autoencoder from a pretrained KL-VAE checkpoint provided by LDM~\cite{rombach2022high}. We keep the pretrained autoencoder weights frozen throughout training. 

\subsection{Auto-Regressive Transformer}
We construct the input sequence for the transformer model from the embeddings \(x_e\) and \(y_e\). During the training phase, the input sequence is constructed by concatenating \(x_e\) with its corresponding \(y_e\), where \(x_e\) remains fully unmasked, while \(y_e\) is selectively masked—either partially or fully. During inference, since \(y_e\) is unavailable, the input sequence is constructed by concatenating \(x_e\) with a zero-valued fully masked token of shape \(L \times Z\), serving as a placeholder for the missing semantic embeddings. The transformer encoder first processes the unmasked sequence with positional embeddings~\cite{vaswani2017attention}. The encoded sequence is then passed through the transformer decoder. The decoder output serves as the positional conditioning variable \( z \) for the subsequent diffusion model to generate the masked semantic embeddings at their respective positions

As a transformer, we adopt widely used autoregressive framework for the sequence modeling architecture. Traditional autoregressive transformers employ causal attention to enforce strict left-to-right dependencies among tokens. However, recent advances, such as Masked Autoencoders (MAE)~\cite{he2022masked}, have demonstrated that bidirectional attention can be effectively utilized for masked sequence prediction. Inspired by this, we integrate bidirectional attention into our transformer. Unlike causal attention, which restricts embedding interactions to preceding elements, our approach enables all visible embeddings to attend to each other while allowing masked embeddings to leverage contextual information.

\textbf{Implementation Details:} We implement ViT transformer~\cite{vaswani2017attention, dosovitskiy2020image} architecture. Both encoder and decoder consist of a stack of 16 Transformer blocks. We employ a random masking strategy during training inspired by masked generative modeling~\cite{chang2022maskgit, li2023mage}. The masking ratio is sampled from a truncated Gaussian distribution centered at 100\% masking with a minimum mask ratio of 70\%.

\subsection{Diffusion Model}
The objective of the diffusion model is to compute the loss for backpropagation during training and facilitate sampling for semantic embedding generation during inference.

\paragraph{Training Phase}
Given \(\mathbf{y}_e\) and the transformer-generated condition vector \(z\), our objective is to model the conditional probability distribution \(p(y_e|z)\) via diffusion. We employ a denoising-based loss function (Eq.~\ref{eq:diffusionloss}), as proposed in~\cite{ho2020denoising, nichol2021improved, dhariwal2021diffusion}.

\begin{equation}
L(z, x) = E_{\epsilon,t} \big[ \|\epsilon - \epsilon_{\theta}({y_e}_t | t, z) \|^2 \big]
\label{eq:diffusionloss}
\end{equation}

where \( \epsilon \sim \mathcal{N}(0, I) \) is a noise sample drawn from a standard normal distribution, and \({y_e}_t=\bar{\alpha}_t y_e + \sqrt{1 - \bar{\alpha}_t} \epsilon\) represents a noise-corrupted version of \(y_e\) at timestep \( t \). Here, \( \bar{\alpha}_t \) defines the noise schedule over time. The function \( \epsilon_{\theta}(x_t | t, z) \) is a learnable noise estimator parameterized by a lightweight MLP network~\cite{ho2020denoising, nichol2021improved}. This model takes the noisy embedding \({y_e}_t\), the conditioning variable \( z \), and the noise schedule timestep \(t\) as input, aiming to estimate the noise component \(\epsilon\).

\paragraph{Inference Phase}
During inference, the objective is to sample from the conditional distribution \(p(y_e|z)\). This is accomplished through a reverse diffusion process, following DDPM~\cite{dhariwal2021diffusion, ho2020denoising}. The iterative denoising procedure is formulated by Eq.\ref{eq:denoising}.
\begin{equation}
{y_e}_{(t-1)} = \frac{1}{\sqrt{\alpha_t}} \left( {y_e}_t - \frac{\sqrt{1 - \alpha_t}}{\sqrt{1 - \bar{\alpha}_t}} \epsilon_{\theta}({y_e}_t | t, z) \right) + \sigma_t \delta
\label{eq:denoising}
\end{equation}

where \( \delta \sim \mathcal{N}(0, I) \) is sampled from a standard Gaussian distribution, and \( \sigma_t \) represents the noise level at timestep \( t \). The process begins with an initial noise sample \(\epsilon\) and iteratively refines it through denoising steps until reaching an approximation of \( p(y_e|z)\).

\textbf{Implementation Details:} We employ a lightweight MLP composed of multiple residual blocks~\cite{he2016deep} as diffusion model. The diffusion process is implemented in accordance with~\cite{nichol2021improved} following a cosine trajectory, utilizing 1000 steps during training. During inference, the process is accelerated by resampling with a reduced steps, set to 100. 

\subsection{Semantic Segmentation Label Reconstruction}
During inference, the output of the diffusion model is decoded into a semantic image \(y_i\) using the decoder part of the autoencoder. To obtain the final segmentation map \(y_m\), \(y_i\) is mapped back to its quantized label space using a nearest-neighbor search. This step is entirely non-parametric, requiring no additional training.

\section{Experiment}
We demonstrate that continuous-valued embeddings outperform quantized embeddings in generating semantic images. To support this, we conduct a performance analysis comparing continuous-valued and quantized embeddings. We also demonstrate that continuous-valued embeddings yield strong segmentation results on the source dataset while exhibiting robust zero-shot domain adaptation across diverse environments using the base transformer (as mentioned earlier our transformer model has only 16 blocks). Finally, we demonstrate noise robustness in continuous-valued embedding through extensive evaluations across various noises.

\paragraph{Dataset} We comprehensively evaluate our model on four benchmark datasets: Cityscapes (mentioned as City in the tables and figures)~\cite{cordts2015cityscapes, cordts2016cityscapes}, SemanticKITTI (mentioned as KITTI in the tables and figures)~\cite{geiger2012we, weber2021step}, ACDC (mentioned as Fog, Rain, Snow in the tables and figures)~\cite{sakaridis2021acdc}, and CADEdgeTune (mentioned as CAD in the tables and figures)~\cite{ahmed2023online, ahmed2024arsfinetune}. The Cityscapes dataset comprises high-resolution images (\(2048\times1024\)) capturing urban street scenes across multiple cities in Germany. To assess zero-shot domain adaptation, we evaluate our model on datasets that share similarities with Cityscapes but introduce domain variations. SemanticKITTI, for example, contains images with different resolutions (\(1242\times375\)) and generally brighter lighting conditions, introducing a domain shift in terms of image scale and illumination. ACDC captures challenging weather conditions such as Fog, Rain, and Snow, allowing us to analyze model robustness under adverse conditions. For a detailed insight, we report separate performance metrics for each weather category. Unlike Cityscapes, which is collected from a well-positioned car, CADEdgeTune is acquired using a small unmanned ground vehicle (UGV). This results in a lower camera viewpoint and semi-urban surroundings, introducing additional variations in perspective and scene composition. By evaluating across these diverse datasets, we provide a comprehensive assessment of our model’s generalization capabilities, particularly in zero-shot domain adaptation and robustness against environmental variations. 

\subsection{Continuous-valued vs Quantized Embedding}
\begin{figure}[!http]
    \centering
    \begin{minipage}{0.23\textwidth}
        \centering
        \includegraphics[width=\linewidth]{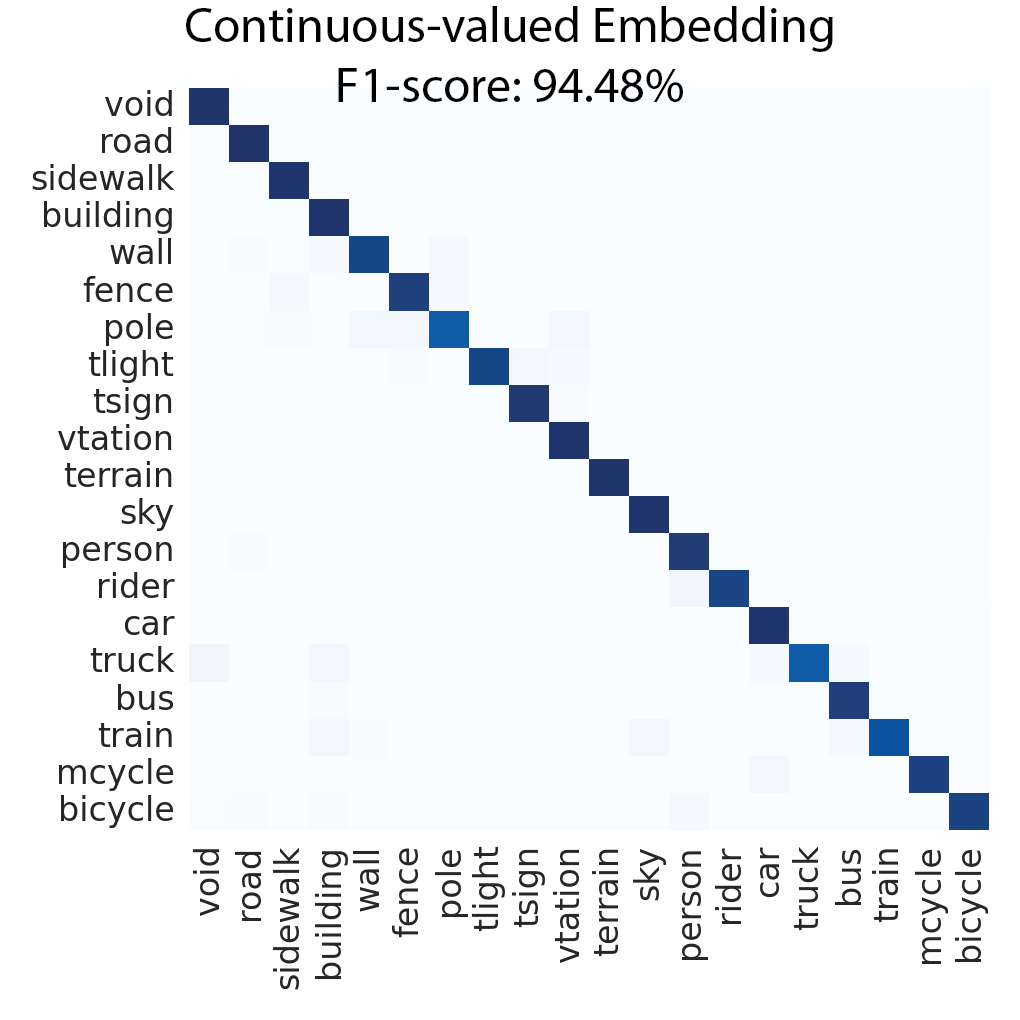}
        \subcaption{KL-VAE}
        \label{fig:klvae}
    \end{minipage}
    \begin{minipage}{0.23\textwidth}
        \centering
        \includegraphics[width=\linewidth]{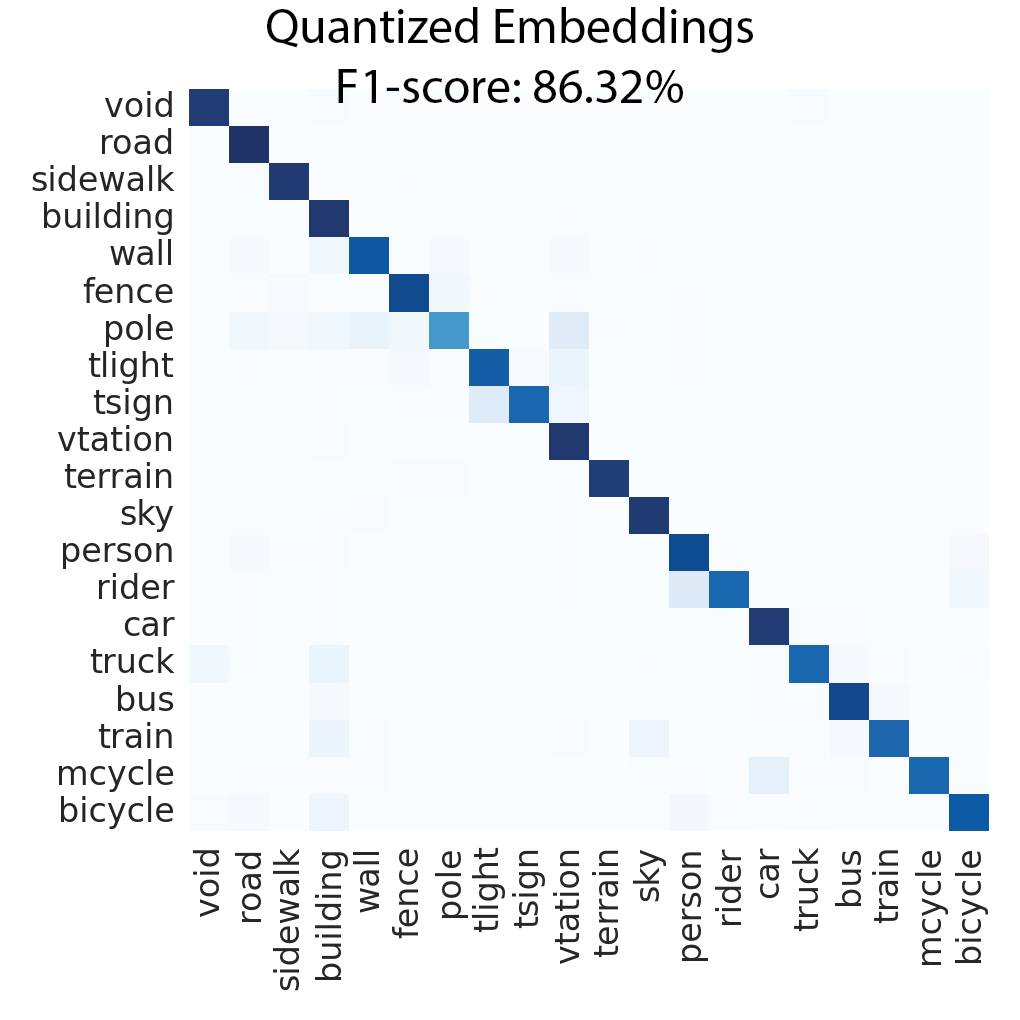}
        \subcaption{VQ-VAE}
        \label{fig:vqvae}
    \end{minipage}
    \caption{Comparison of segmentation image reconstruction between Continuous-valued Embedding (KL-VAE) and Quantized Embedding (VQ-VAE).}
    \label{fig:cm}
\end{figure}

\paragraph*{\textbf{Approach}} We experiment with the KL-VAE and VQ-VAE models from LDM to compare the performance of continuous-valued and quantized embeddings. We leverage VAE with Imagenet dataset pre-trained weights provided by LDM~\cite{rombach2022high} without fine-tuning. This allows us to determine the direct applicability of these models to a broader range of tasks without additional training. Since we focus on the semantic segmentation application, we evaluate performance based on semantic image reconstruction accuracy rather than RGB image reconstruction.
\paragraph*{\textbf{Result}} Fig.~\ref{fig:cm} presents the confusion matrices for the KL-VAE and VQ-VAE models, highlighting the misclassification patterns in their segmentation predictions.
Our quantitative analysis demonstrates that continuous-valued embeddings (KL-VAE) significantly outperform quantized embeddings (VQ-VAE) in semantic segmentation tasks. The key findings are as follows.
\begin{itemize}
    \item \textbf{Continuous-valued embeddings (KL-VAE):} KL-VAE achieves superior segmentation performance with an F1 score of 94.48\%- consistently excelling across all object categories, as shown in Fig.~\ref{fig:klvae}. Its continuous-valued embeddings capture fine-grained details more effectively.
    \item \textbf{Quantized Embeddings (VQ-VAE):} VQ-VAE, with an F1 score of 86.32\%- struggles with rare and small objects such as motorcycles, bicycles, and traffic lights, as illustrated in Fig.~\ref{fig:vqvae}. The inherent information loss from vector quantization leads to higher misclassification rates.
\end{itemize}

\subsection{Semantic Segmentation Analysis}
\paragraph*{\textbf{Approach}} We train CAM-Seg using the Cityscapes dataset, while SemanticKITTI, ACDC, and CADEdgeTune are excluded from training to evaluate zero-shot domain adaptation. We report standard Precision and Average Precision (AP) metrics as segmentation performance. Additionally, following the SAM approach~\cite{kirillov2023segment}, we analyze segmentation performance based on object size and occurrence frequency to understand the model’s behavior. For size-based categorization, we define small objects as those with fine structures, including poles, traffic lights, traffic signs, persons, and riders. Medium-sized objects include fences, cars, trucks, buses, trains, motorcycles, and bicycles, while large objects comprise roads, sidewalks, buildings, vegetation, terrain, sky, and walls. Similarly, for frequency-based categorization, frequent objects such as roads, vegetation, buildings, sky, and sidewalks appear consistently across the dataset. Common objects include cars, terrain, walls, fences, persons, poles, and trucks, while rare objects—such as buses, traffic signs, bicycles, riders, traffic lights, trains, and motorcycles—are less frequently encountered. This categorization enables a fine-grained evaluation of segmentation performance across diverse object types.

\begin{figure*}[!http]
\centering
\includegraphics[width=\linewidth]{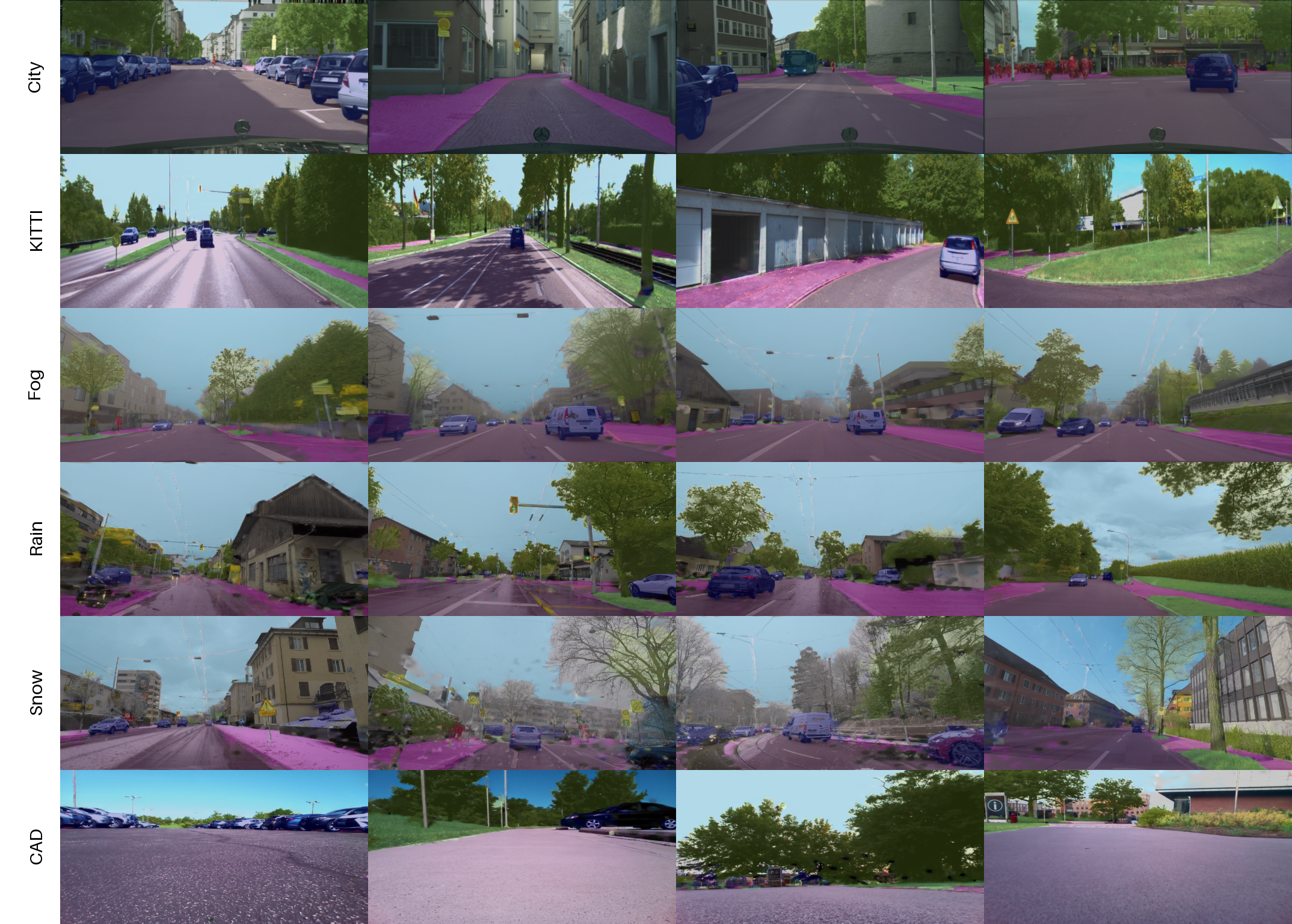}
\caption{Qualitative segmentation results across different datasets, illustrating the model's performance on diverse environmental conditions including urban (City), different perspectives (KITTI, CAD), and adverse weather scenarios (Fog, Rain, Snow).}
\label{fig:qualitative}
\end{figure*}

\paragraph*{\textbf{Results}} Tab.~\ref{tab:resultsummary} provides a high-level overview of segmentation performance across different datasets, while Tab.~\ref{tab:resultdetail} presents a more granular breakdown across individual object classes.  Furthermore, qualitative results, which show the visual outputs of semantic segmentation, are illustrated in Fig.~\ref{fig:qualitative}. The key findings from this result are as follows.
\begin{table}[!http]
\caption{Summarized segmentation performance across different datasets, reporting overall Average Precision (AP) along with size-based (small, medium, large) and frequency-based (frequent, common, rare) category-wise AP scores.}
\label{tab:resultsummary}
\centering
\resizebox{\columnwidth}{!}{
\begin{tabular}{lc|ccc|ccc}
Dataset & AP & AP\textsuperscript{S} & AP\textsuperscript{M} & AP\textsuperscript{L} & AP\textsuperscript{F} & AP\textsuperscript{C} & AP\textsuperscript{R}\\
\hline
\hline
City & 70.23 &  68.16 &  57.67 &  84.26 &  92.46 &  63.00 &  61.58\\
KITTI & 49.49 &  45.23 &  26.49 &  75.54 &  85.82 &  43.02 &  30.01\\
Fog & 51.99 &  42.94 &  32.11 &  78.33 &  88.18 &  47.96 &  30.17\\
Rain & 43.50 &  32.71 &  20.79 &  73.92 &  86.77 &  34.21 &  21.89\\
Snow & 40.69 &  26.38 &  20.20 &  71.40 &  81.37 &  32.98 &  19.34\\
CAD & 34.89 & 9.13 &  14.44 &  57.63 &  64.54 &  23.10 &   2.08\\
\end{tabular}}
\end{table}
\begin{table*}[!t]
\caption{Detailed per-class segmentation performance across various datasets, highlighting the model’s zero-shot adaptation capabilities to different image resolutions (KITTI), adverse weather conditions (Fog, Rain, Snow), and varying camera viewpoints (CAD).}
\label{tab:resultdetail}
\centering
\vspace{5ex}
\resizebox{2\columnwidth}{!}{
\begin{tabular}{lccccccccccccccccccc}
Dataset                                 & \begin{rotate}{90} Road\end{rotate}  & \begin{rotate}{90}Sidewalk\end{rotate}       & \begin{rotate}{90}Building\end{rotate} & \begin{rotate}{90}Wall\end{rotate}           & \begin{rotate}{90}Fence\end{rotate}          & \begin{rotate}{90}Pole\end{rotate}           & \begin{rotate}{90} TLight\end{rotate}  & \begin{rotate}{90} TSignal\end{rotate} & \begin{rotate}{90}Vtation\end{rotate} & \begin{rotate}{90}Terrain\end{rotate} & \begin{rotate}{90}Sky\end{rotate}   & \begin{rotate}{90}Person\end{rotate} & \begin{rotate}{90}Rider\end{rotate} & \begin{rotate}{90}Car\end{rotate}            & \begin{rotate}{90}Truck\end{rotate}          & \begin{rotate}{90}Bus\end{rotate}            & \begin{rotate}{90}Train\end{rotate}          & \begin{rotate}{90}Mcycle\end{rotate}    & \begin{rotate}{90}Bicycle\end{rotate}   \\
\hline
\hline
City & 98.1 & 86.4 & 89.2 & 47.3 & 43.4 & 60.1 & 63.0 & 82.5 & 92.7 & 80.3 & 96.0 & 70.9 & 64.3 & 94.0 & 45.0 & 66.6 & 43.7 & 48.3 & 62.7\\
\hline
\multicolumn{20}{l}{\textbf{Zero-shot Adaptation to Varying Image Resolutions}}\\
KITTI & 84.3 & 74.4 & 79.4 & 11.9 & 19.5 & 33.0 & 66.8 & 50.2 & 95.1 & 87.8 & 95.9 & 53.1 & 23.0 & 83.3 & 12.5 & 16.6 & 2.2 & 26.7 & 24.6\\
\hline
\multicolumn{20}{l}{\textbf{Zero-shot Adaptation to Adverse Weather Conditions}}\\
Fog & 95.2 & 86.0 & 69.7 & 33.8 & 38.4 & 32.0 & 68.3 & 24.2 & 91.5 & 73.6 & 98.5 & 40.2 & 50.0 & 76.3 & 41.4 & 35.1 & 21.1 & 1.7 & 10.9\\
Rain & 
89.3 & 74.6 & 78.7 & 15.6 & 26.3 & 28.4 & 53.5 & 24.5 & 92.5 & 68.0 & 98.8 & 30.6 & 26.6 & 66.2 & 4.4 & 35.3 & 3.5 & 3.7 & 6.2\\
Snow & 89.4 & 74.1 & 53.1 & 29.0 & 21.4 & 20.3 & 49.5 & 28.1 & 92.3 & 64.0 & 97.9 & 29.5 & 4.5 & 60.0 & 6.8 & 1.9 & 48.1 & 1.1 & 2.2\\
\hline
\multicolumn{20}{l}{\textbf{Zero-shot Adaptation to Varying Camera Viewpoints}}\\
CAD & 40.2 & 63.8 & 35.4 & 1.8 & 20.4 & 16.5 & \textemdash{} & 4.0 & 87.5 & 78.9 & 95.7 & 6.8 & \textemdash{} & 36.8 & 0.4 & 0.1 & \textemdash{} & \textemdash{} & \textemdash{}\\
\end{tabular}}
\end{table*}

\begin{itemize}
     \item \textbf{Performance on Frequent and Large Objects:} As shown in Tab~\ref{tab:resultsummary}, the model demonstrates strong performance on frequent and large object classes such as roads, vegetation, and sky. Even when evaluated on unseen datasets, the precision remains at an acceptable level, highlighting the model’s ability to generalize to common structural elements in diverse environments.
    
    \item \textbf{Challenges with Rare Objects:} As observed in Tab~\ref{tab:resultsummary}, the precision for rare object classes is comparatively lower. However, the model still successfully detects these objects, indicating its capability to recognize less frequent categories despite the reduced training exposure.
    
    \item \textbf{Impact of Weather Conditions:} Segmentation performance is more resilient to fog compared to heavy rain and snow. As presented in Tab~\ref{tab:resultdetail}, the model maintains acceptable precision for most classes except for trucks, motorcycles, and bicycles, where adverse weather significantly affects visibility.
    
    \item \textbf{Camera Viewpoint Sensitivity:} The model exhibits slightly lower segmentation accuracy on the CADEdgeTune dataset than others. As illustrated in Fig.~\ref{fig:qualitative}, drastic changes in camera angles and variations in urban layouts contribute to misclassification, particularly in large structural elements such as buildings and roads.
\end{itemize}

To the best of our knowledge, there are no existing works that apply continuous-valued embedding based autoregressive transformers for semantic segmentation. Due to the absence of directly comparable approaches, we are unable to report a benchmark comparison. Instead, our results serve as a first-of-its-kind evaluation, providing a baseline for future research in continuous-valued embedding modeling for semantic segmentation.

\begin{figure*}[!t]
    \centering
    \begin{minipage}{0.24\textwidth}
        \centering
        \includegraphics[width=\linewidth]{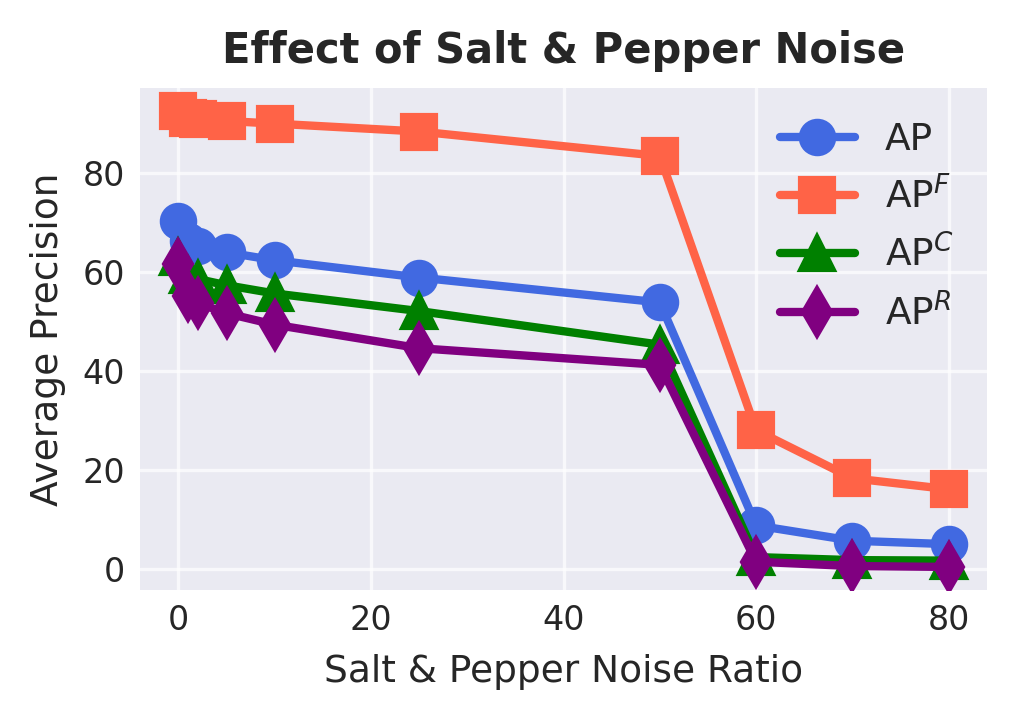}
        \subcaption{Salt \& Pepper Noise}
        \label{fig:noisesalt}
    \end{minipage}
    \begin{minipage}{0.24\textwidth}
        \centering
        \includegraphics[width=\linewidth]{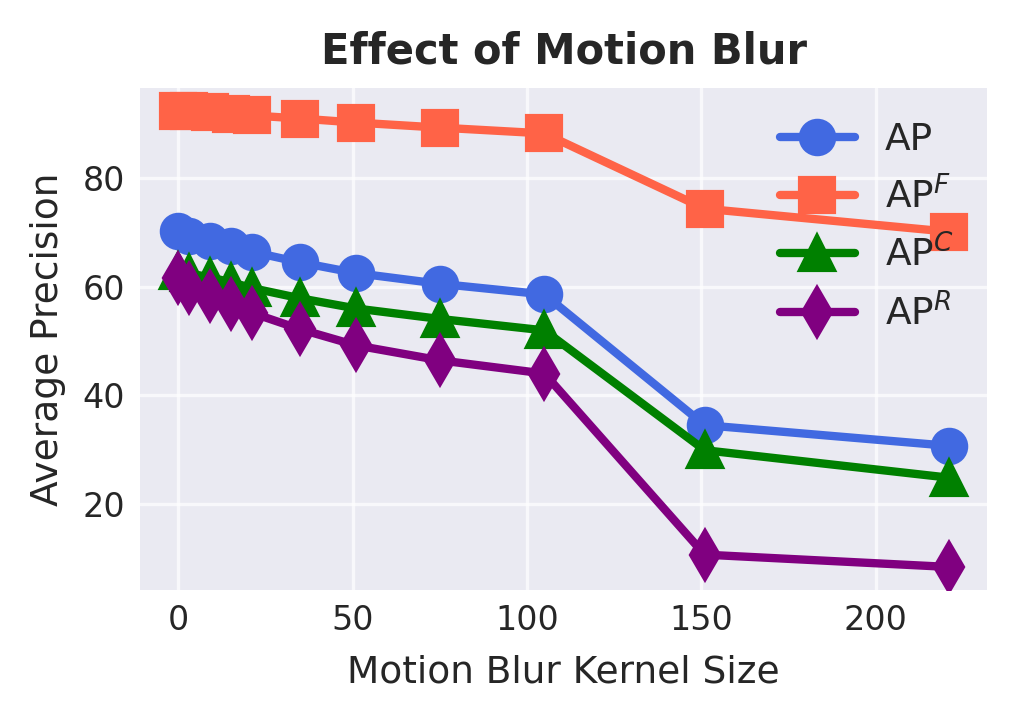}
        \subcaption{Motion Blur}
        \label{fig:noisemblur}
    \end{minipage}
    \begin{minipage}{0.24\textwidth}
        \centering
        \includegraphics[width=\linewidth]{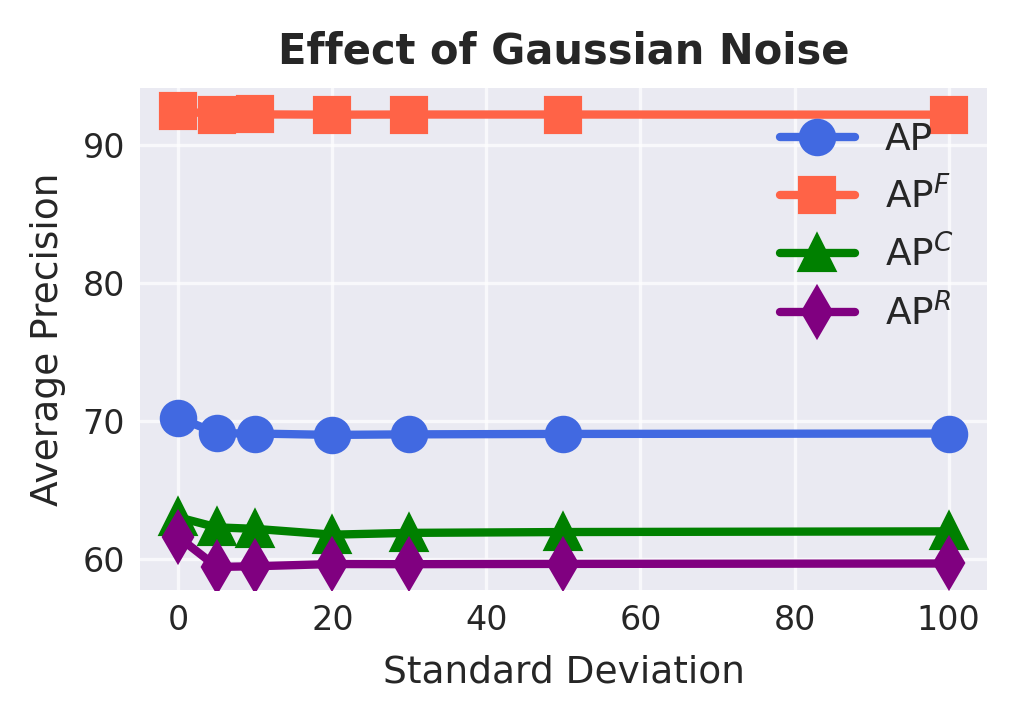}
        \subcaption{Gaussian Noise}
        \label{fig:noisegaussian}
    \end{minipage}
    \begin{minipage}{0.24\textwidth}
        \centering
        \includegraphics[width=\linewidth]{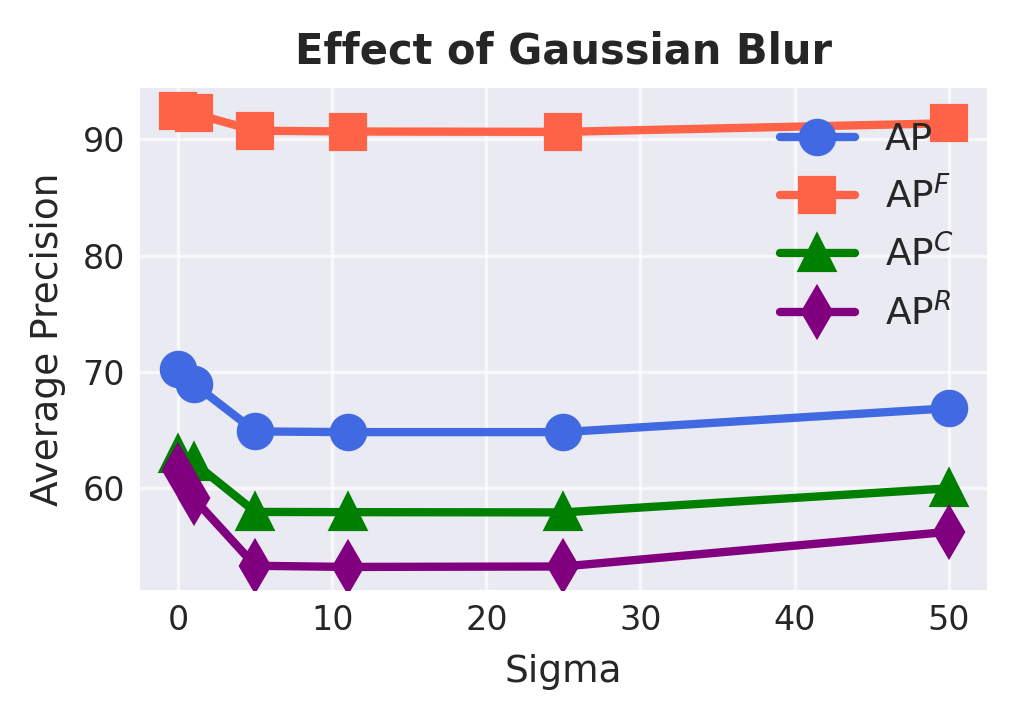}
        \subcaption{Gaussian Blur}
        \label{fig:noisegblur}
    \end{minipage}


    \begin{minipage}{0.24\textwidth}
        \centering
        \includegraphics[width=\linewidth]{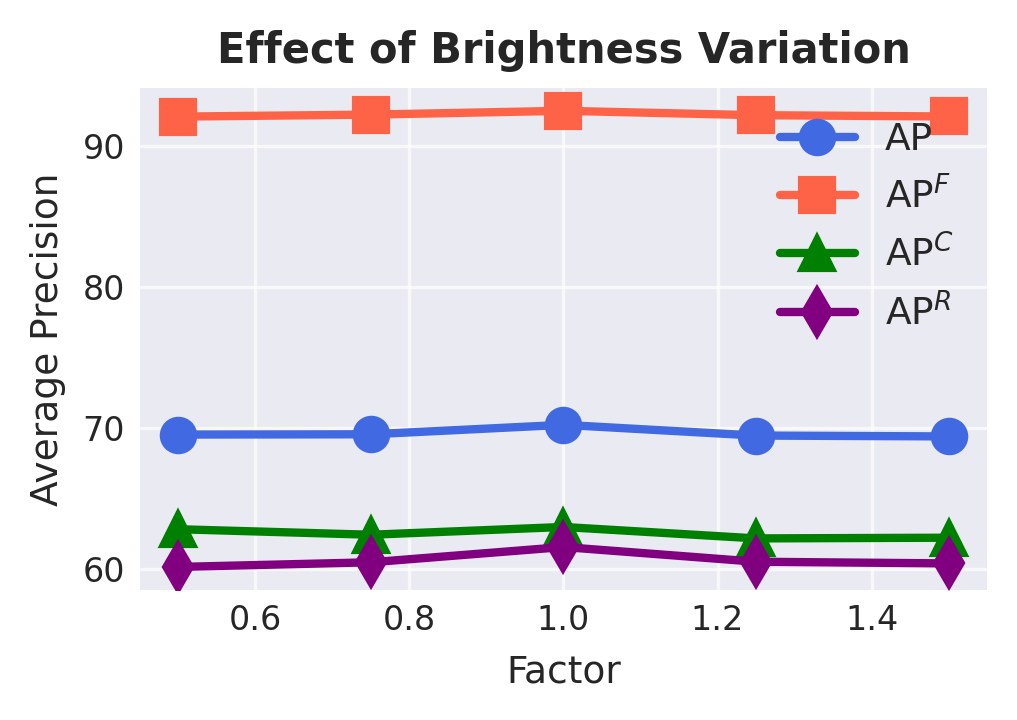}
        \subcaption{Brightness Variation}
        \label{fig:noisebirghtness}
    \end{minipage}
    \begin{minipage}{0.24\textwidth}
        \centering
        \includegraphics[width=\linewidth]{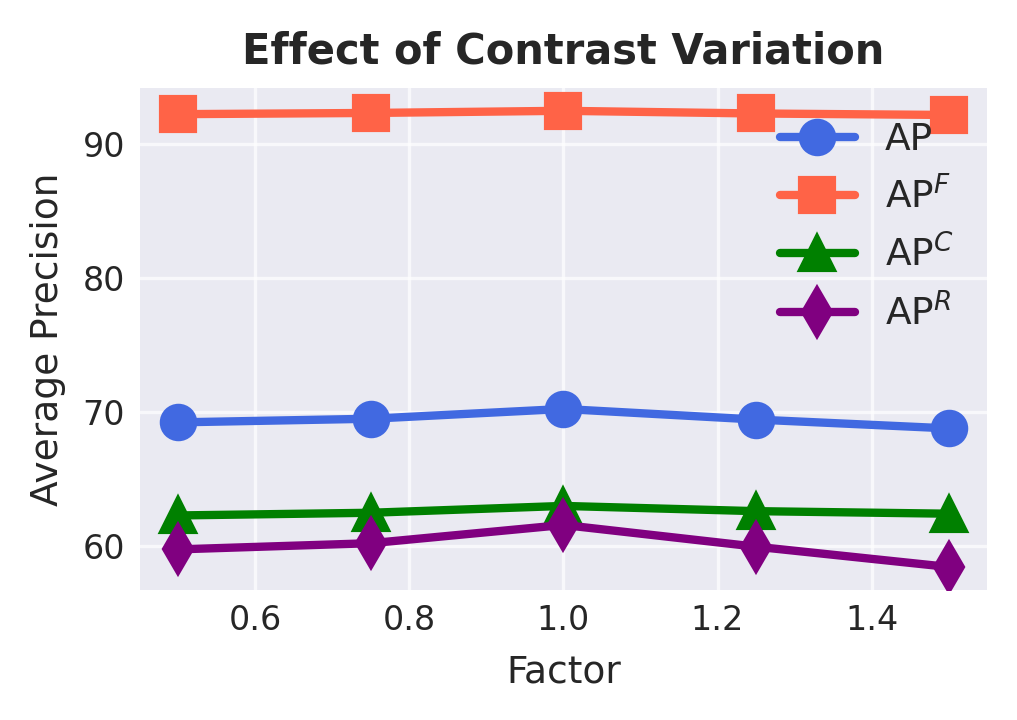}
        \subcaption{Contrast Variation}
        \label{fig:noisecontrass}
    \end{minipage}
    \begin{minipage}{0.24\textwidth}
        \centering
        \includegraphics[width=\linewidth]{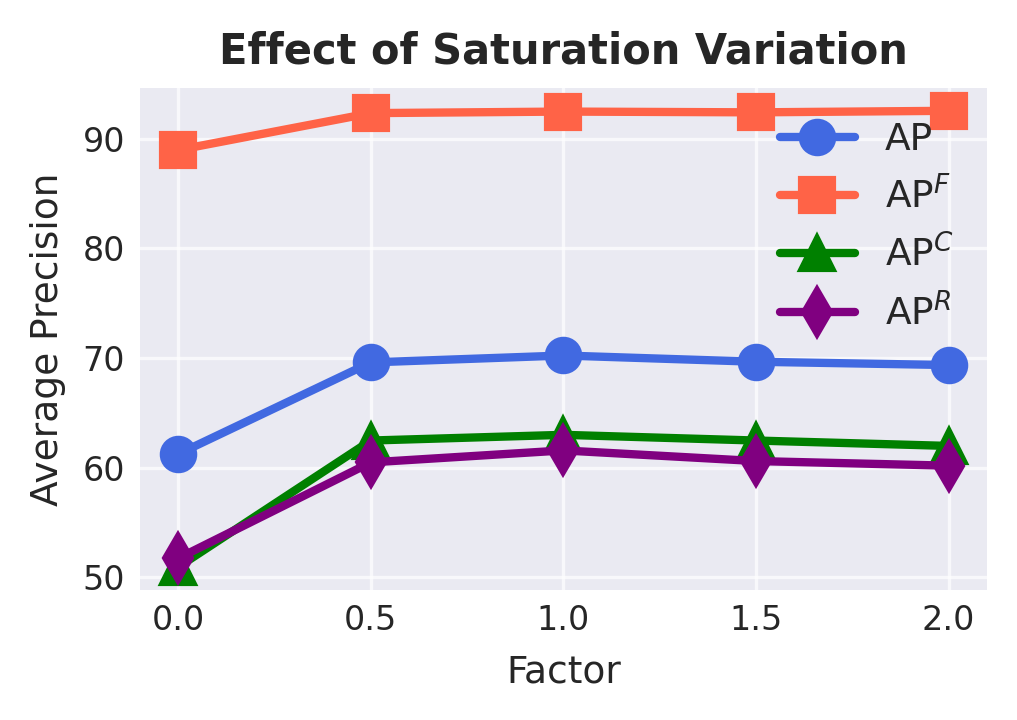}
        \subcaption{Saturation Variation}
        \label{fig:noisesaturation}
    \end{minipage}
    \begin{minipage}{0.24\textwidth}
        \centering
        \includegraphics[width=\linewidth]{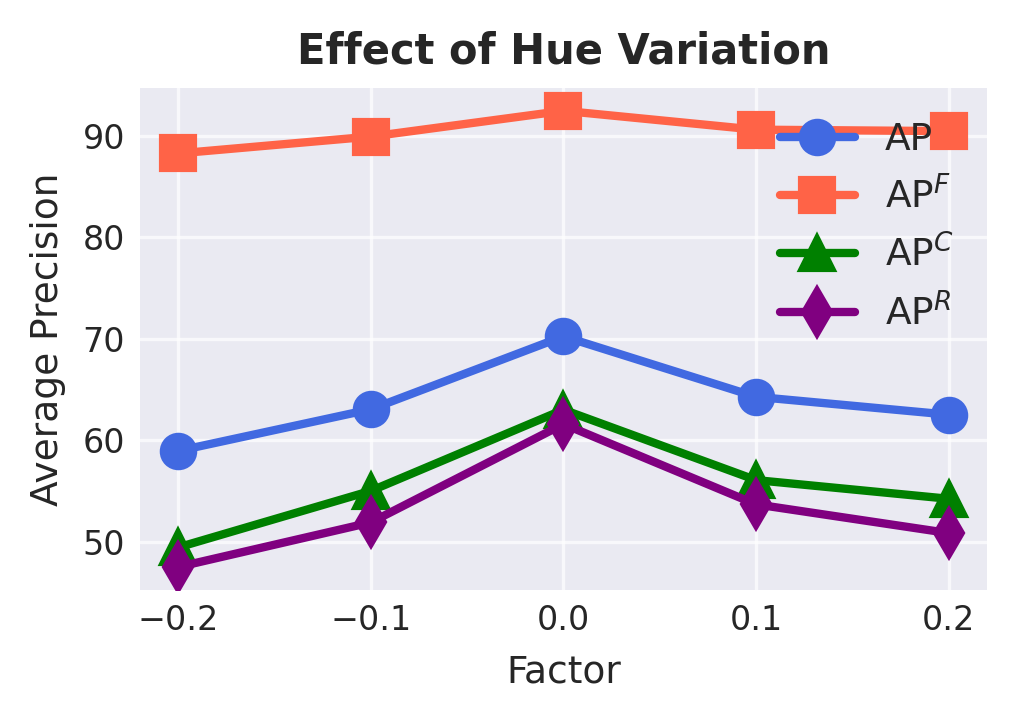}
        \subcaption{Hue Variation}
        \label{fig:noisehue}
    \end{minipage}
    \caption{Quantitative results of semantic segmentation under various noise conditions.}
    \label{fig:noise}
\end{figure*}
\subsection{Effects of Noise}
\paragraph*{\textbf{Approach}} To assess the robustness and generalization under degraded conditions, we introduce various noise perturbations in RGB image space and evaluate their impact on semantic segmentation performance. We experiment with different types of noise distortions in real-world scenarios. Salt \& pepper noise is applied at varying levels from 1\% to 80\% to simulate sensor corruption and extreme compression artifacts. Motion blur is introduced with kernel sizes ranging from 3 to 151, mimicking different levels of camera shake and object motion. Gaussian noise, with a mean of zero and a standard deviation between 5 and 100, is added to simulate electronic sensor noise. Gaussian blur is applied using a fixed kernel size of 11 while varying the standard deviation from 1 to 50 to replicate defocus or atmospheric distortions. Lastly, color jitter is used to adjust brightness (0.5 to 1.5 scale), contrast (0.5 to 1.5 scale), saturation (0 to 2 scale), and hue (-0.2 to 0.2 scale), mimicking variations in lighting, sensor inconsistencies, and environmental illumination shifts. This evaluation demonstrates the model’s ability to maintain segmentation accuracy under various challenging conditions.

\paragraph*{\textbf{Results}} Fig.~\ref{fig:noise} illustrates the quantitative impact of various noise perturbations on the performance of the semantic segmentation. From the quantitative results, we derive the following observations. 

\begin{itemize}
    \item \textbf{Salt \& Pepper Noise:} The model maintains stable performance up to 50\% noise, but beyond this threshold, segmentation quality deteriorates, particularly for rare and small objects (Fig.~\ref{fig:noisesalt}).
    \item \textbf{Motion Blur:} Mild blur (kernel size $\leq 35$) has minimal impact on AP scores, while extreme blur (kernel size $\geq 151$) results in a significant decline (Fig.~\ref{fig:noisemblur}).
    \item \textbf{Gaussian Noise:} The model exhibits strong resilience, with less than a 2\% drop in AP even at extreme noise levels ($\sigma = 100$) (Fig.~\ref{fig:noisegaussian}).
    \item \textbf{Gaussian Blur:} Low to moderate blur ($\sigma \leq 25$) has minimal effects, primarily degrading edge clarity. Interestingly, higher blur levels ($\sigma = 50$) lead to slight performance recovery, likely due to smoothed object representations (Fig.~\ref{fig:noisegblur}).
    \item \textbf{Brightness Variation:} Minor brightness shifts (0.75–1.25 scale) have negligible impact, whereas extreme adjustments (0.5, 1.5 scale) slightly degrade performance (Fig.~\ref{fig:noisebirghtness}).
    \item \textbf{Contrast Variation:} Increasing contrast beyond a 1.5 scale reduces segmentation performance as higher contrast alters object boundary visibility (Fig.~\ref{fig:noisecontrass}).
    \item \textbf{Saturation Variation:} Desaturation (factor = 0) significantly reduces segmentation quality, while increased saturation (factor = 2) causes minor fluctuations (Fig.~\ref{fig:noisesaturation}).
    \item \textbf{Hue Variation:} Hue shifts degrade segmentation accuracy, leading to misclassification of objects with similar colors (Fig.~\ref{fig:noisehue}).    
\end{itemize}

\section{Ablation Studies}
In this section, we analyze key design factors that influence the performance of our framework. Specifically, we assess how input image size, and the choice of semantic image color mapping influence the model’s ability to generate accurate and robust segmentation maps.

\subsection{Impact of Input Image Size}
\paragraph*{\textbf{Approach}} To evaluate the impact of input image size on segmentation performance, we train our model using different input resolutions, analyzing how the model captures spatial dependencies and generalizes across datasets. We experiment with three different resolutions: \(256\times 256\), which results in a context length of 256; \(512\times 512\), yielding a context length of 1024; and \(768\times 768\), producing a context length of 2304. 

\begin{figure}[!http]
\centering
\includegraphics[width=.9\linewidth]{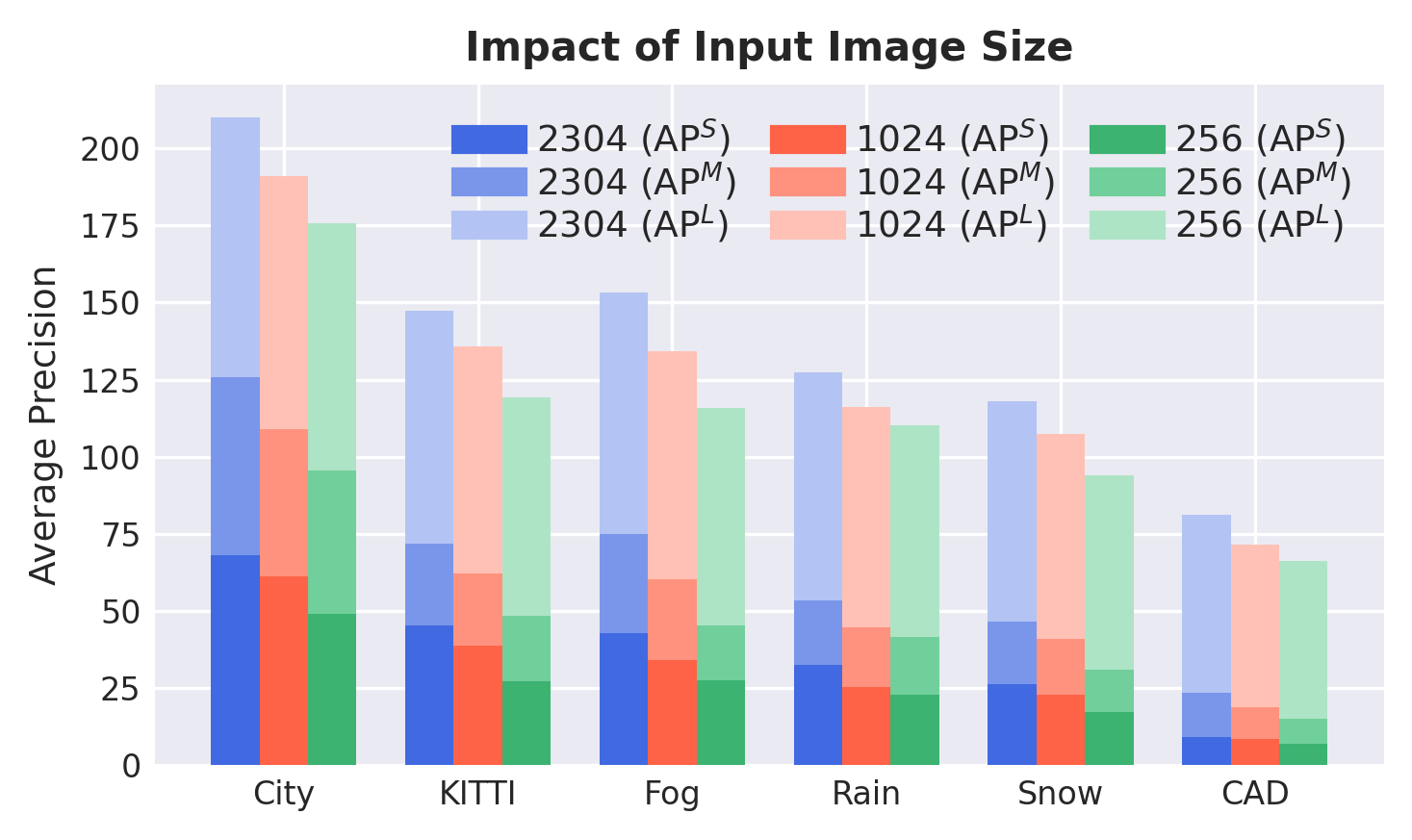}
\caption{Impact of input image size on segmentation.}
\label{fig:context_len}
\end{figure}

\paragraph*{\textbf{Result}} 
Fig.~\ref{fig:context_len} illustrates segmentation performance across datasets for different input image sizes.
Our analysis reveals that increasing the input image size consistently improves segmentation quality, particularly for medium and small objects. As shown in Fig.~\ref{fig:context_len}, models trained with larger image resolutions (\(768 \times 768\), context length: 2304), highlighted by blue bars, achieve the highest segmentation accuracy. This demonstrates that a broader spatial context enhances feature extraction and preserves fine-grained details. Conversely, models trained with smaller resolutions (\(256 \times 256\), context length: 256), represented by green bars, exhibit noticeable performance degradation. This suggests that a limited spatial context restricts the model’s ability to capture intricate structures effectively. Furthermore, as shown in the third, fourth, and fifth columns of Fig.~\ref{fig:context_len}, models trained with higher-resolution inputs demonstrate superior robustness in adverse conditions such as fog, rain, and snow. This suggests that longer spatial context improves the model's resilience to domain shifts, enabling more accurate segmentation in challenging environments.

\subsection{Impact of Semantic Image Color Palette}
\begin{figure}[!http]
\centering
\includegraphics[width=\linewidth]{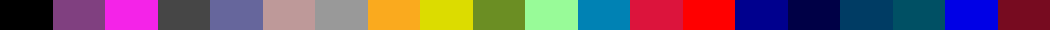}
\includegraphics[width=\linewidth]{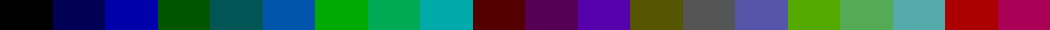}
\caption{Semantic segmentation image color palettes.}
\label{fig:palletecolor}
\end{figure}

\paragraph*{\textbf{Approach}} To assess whether the choice of color palette in semantic images influences model performance, we evaluate the accuracy of an autoencoder trained on the Cityscapes dataset with different color palettes. Specifically, we compare the standard Cityscapes palette~\cite{cordts2016cityscapes} against an inpainting-based palette~\cite{bar2022visual} to determine if color variations affect segmentation quality. 
\paragraph*{\textbf{Result}} The comparison color using the Cityscapes color palette and the inpainting-based color palette is illustrated in Fig.~\ref{fig:palletecolor}.
Despite the difference in color mapping, the F1-scores remain nearly identical, with the Cityscapes palette achieving 94.48\% and the inpainting-based palette scoring 93.44\%. This minimal variation suggests that the semantic segmentation model is largely invariant to color palette differences, implying that the underlying learned representations are robust to color-based variations. Consequently, the choice of a specific color palette does not significantly impact semantic segmentation accuracy.

\section{Discussion}
Despite being trained on a relatively small dataset and a lightweight transformer model, our approach demonstrates strong performance on the source data distribution and zero-shot domain adaptation across multiple unseen datasets. This highlights the potential of continuous-valued embeddings in enhancing autoregressive models for vision tasks. Our results suggest that transitioning from quantized to continuous-valued embeddings could lead to more expressive and generalizable representations for semantic segmentation. However, our model struggles in extremely dark environments, indicating the need for further improvements in robustness under low-light conditions.

For future work, we aim to scale our approach by training on a larger and more diverse dataset encompassing a wide range of environmental conditions beyond road-side scenarios. Additionally, we plan to integrate a larger transformer model to further enhance segmentation performance. Another key direction is incorporating cross-talk mechanisms between our model and other vision models, such as SAM~\cite{kirillov2023segment}, SegFormer~\cite{xie2021segformer}, and SegGPT~\cite{wang2023seggpt}, leveraging their encoder and decoder structures to enrich feature representations. Finally, we envision developing a large-scale vision model similar to LVM~\cite{bai2024sequential}, capable of performing multiple vision tasks, including object detection and depth estimation, using a unified prompt-based framework.

\section{Conclusion}
In this work, we introduce CAM-Seg, a novel approach for RGB image-conditioned semantic image generation in semantic segmentation applications, leveraging a continuous-valued feature space. Our method employs an autoregressive transformer combined with a diffusion model, trained using diffusion loss, to directly map RGB image space embeddings to semantic image space embeddings—eliminating the need for vector quantization. Extensive experiments on four real-world datasets demonstrate that our approach preserves richer feature representations, enhancing segmentation accuracy, robustness under diverse noise, zero-shot adaptation, adverse weather conditions, and varying viewpoints. Ablation studies validate the model's generalization capability and flexibility, underscoring its potential as a robust and scalable solution for semantic image generation.

\section{Acknowledgement}
This work has been partially supported by U.S. Army Grant \#W911NF2120076, U.S. Army Grant \#W911NF2410367, ONR Grant \#N00014-23-1-2119, NSF CAREER Award \#1750936, NSF REU Site Grant \#2050999, and NSF CNS EAGER Grant \#2233879.

{
    \small
    \bibliographystyle{ieeenat_fullname}
    \bibliography{main}

\begin{thebibliography}{50}
\providecommand{\natexlab}[1]{#1}
\providecommand{\url}[1]{\texttt{#1}}
\expandafter\ifx\csname urlstyle\endcsname\relax
  \providecommand{\doi}[1]{doi: #1}\else
  \providecommand{\doi}{doi: \begingroup \urlstyle{rm}\Url}\fi

\bibitem[Ahmed et~al.(2023)Ahmed, Hasan, Yingling, O’Leary, Purushotham, You, and Roy]{ahmed2023online}
Masud Ahmed, Zahid Hasan, Tim Yingling, Eric O’Leary, Sanjay Purushotham, Suya You, and Nirmalya Roy.
\newblock An online continuous semantic segmentation framework with minimal labeling efforts.
\newblock In \emph{2023 IEEE International Conference on Smart Computing (SMARTCOMP)}, pages 116--123. IEEE, 2023.

\bibitem[Ahmed et~al.(2024)Ahmed, Hasan, Faridee, Anwar, Jayarajah, Purushotham, You, and Roy]{ahmed2024arsfinetune}
Masud Ahmed, Zahid Hasan, Abu Zaher~Md Faridee, Mohammad~Saeid Anwar, Kasthuri Jayarajah, Sanjay Purushotham, Suya You, and Nirmalya Roy.
\newblock Arsfinetune: On-the-fly tuning of vision models for unmanned ground vehicles.
\newblock In \emph{2024 20th International Conference on Distributed Computing in Smart Systems and the Internet of Things (DCOSS-IoT)}, pages 170--178. IEEE, 2024.

\bibitem[Bai et~al.(2024)Bai, Geng, Mangalam, Bar, Yuille, Darrell, Malik, and Efros]{bai2024sequential}
Yutong Bai, Xinyang Geng, Karttikeya Mangalam, Amir Bar, Alan~L Yuille, Trevor Darrell, Jitendra Malik, and Alexei~A Efros.
\newblock Sequential modeling enables scalable learning for large vision models.
\newblock In \emph{Proceedings of the IEEE/CVF Conference on Computer Vision and Pattern Recognition}, pages 22861--22872, 2024.

\bibitem[Bar et~al.(2022)Bar, Gandelsman, Darrell, Globerson, and Efros]{bar2022visual}
Amir Bar, Yossi Gandelsman, Trevor Darrell, Amir Globerson, and Alexei Efros.
\newblock Visual prompting via image inpainting.
\newblock \emph{Advances in Neural Information Processing Systems}, 35:\penalty0 25005--25017, 2022.

\bibitem[Caron et~al.(2021)Caron, Touvron, Misra, J{\'e}gou, Mairal, Bojanowski, and Joulin]{caron2021emerging}
Mathilde Caron, Hugo Touvron, Ishan Misra, Herv{\'e} J{\'e}gou, Julien Mairal, Piotr Bojanowski, and Armand Joulin.
\newblock Emerging properties in self-supervised vision transformers.
\newblock In \emph{Proceedings of the IEEE/CVF international conference on computer vision}, pages 9650--9660, 2021.

\bibitem[Chang et~al.(2022)Chang, Zhang, Jiang, Liu, and Freeman]{chang2022maskgit}
Huiwen Chang, Han Zhang, Lu Jiang, Ce Liu, and William~T Freeman.
\newblock Maskgit: Masked generative image transformer.
\newblock In \emph{Proceedings of the IEEE/CVF Conference on Computer Vision and Pattern Recognition}, pages 11315--11325, 2022.

\bibitem[Chen et~al.(2023)Chen, Lu, Zhu, and Zhang]{chen2023generative}
Jiaqi Chen, Jiachen Lu, Xiatian Zhu, and Li Zhang.
\newblock Generative semantic segmentation.
\newblock In \emph{Proceedings of the IEEE/CVF conference on computer vision and pattern recognition}, pages 7111--7120, 2023.

\bibitem[Chen et~al.(2020)Chen, Radford, Child, Wu, Jun, Luan, and Sutskever]{chen2020generative}
Mark Chen, Alec Radford, Rewon Child, Jeffrey Wu, Heewoo Jun, David Luan, and Ilya Sutskever.
\newblock Generative pretraining from pixels.
\newblock In \emph{International conference on machine learning}, pages 1691--1703. PMLR, 2020.

\bibitem[Cheng et~al.(2021)Cheng, Schwing, and Kirillov]{cheng2021per}
Bowen Cheng, Alex Schwing, and Alexander Kirillov.
\newblock Per-pixel classification is not all you need for semantic segmentation.
\newblock \emph{Advances in neural information processing systems}, 34:\penalty0 17864--17875, 2021.

\bibitem[Cheng et~al.(2022)Cheng, Misra, Schwing, Kirillov, and Girdhar]{cheng2022masked}
Bowen Cheng, Ishan Misra, Alexander~G Schwing, Alexander Kirillov, and Rohit Girdhar.
\newblock Masked-attention mask transformer for universal image segmentation.
\newblock In \emph{Proceedings of the IEEE/CVF conference on computer vision and pattern recognition}, pages 1290--1299, 2022.

\bibitem[Cordts et~al.(2015)Cordts, Omran, Ramos, Scharw{\"a}chter, Enzweiler, Benenson, Franke, Roth, and Schiele]{cordts2015cityscapes}
Marius Cordts, Mohamed Omran, Sebastian Ramos, Timo Scharw{\"a}chter, Markus Enzweiler, Rodrigo Benenson, Uwe Franke, Stefan Roth, and Bernt Schiele.
\newblock The cityscapes dataset.
\newblock In \emph{CVPR Workshop on the Future of Datasets in Vision}, page~1, 2015.

\bibitem[Cordts et~al.(2016)Cordts, Omran, Ramos, Rehfeld, Enzweiler, Benenson, Franke, Roth, and Schiele]{cordts2016cityscapes}
Marius Cordts, Mohamed Omran, Sebastian Ramos, Timo Rehfeld, Markus Enzweiler, Rodrigo Benenson, Uwe Franke, Stefan Roth, and Bernt Schiele.
\newblock The cityscapes dataset for semantic urban scene understanding.
\newblock In \emph{Proceedings of the IEEE conference on computer vision and pattern recognition}, pages 3213--3223, 2016.

\bibitem[Dhariwal and Nichol(2021)]{dhariwal2021diffusion}
Prafulla Dhariwal and Alexander Nichol.
\newblock Diffusion models beat gans on image synthesis.
\newblock \emph{Advances in neural information processing systems}, 34:\penalty0 8780--8794, 2021.

\bibitem[Dosovitskiy(2020)]{dosovitskiy2020image}
Alexey Dosovitskiy.
\newblock An image is worth 16x16 words: Transformers for image recognition at scale.
\newblock \emph{arXiv preprint arXiv:2010.11929}, 2020.

\bibitem[Esser et~al.(2021)Esser, Rombach, and Ommer]{esser2021taming}
Patrick Esser, Robin Rombach, and Bjorn Ommer.
\newblock Taming transformers for high-resolution image synthesis.
\newblock In \emph{Proceedings of the IEEE/CVF conference on computer vision and pattern recognition}, pages 12873--12883, 2021.

\bibitem[Geiger et~al.(2012)Geiger, Lenz, and Urtasun]{geiger2012we}
Andreas Geiger, Philip Lenz, and Raquel Urtasun.
\newblock Are we ready for autonomous driving? the kitti vision benchmark suite.
\newblock In \emph{2012 IEEE conference on computer vision and pattern recognition}, pages 3354--3361. IEEE, 2012.

\bibitem[Gu et~al.(2024)Gu, Wang, Zhang, Zhang, Zhang, Jaitly, Susskind, and Zhai]{gu2024dart}
Jiatao Gu, Yuyang Wang, Yizhe Zhang, Qihang Zhang, Dinghuai Zhang, Navdeep Jaitly, Josh Susskind, and Shuangfei Zhai.
\newblock Dart: Denoising autoregressive transformer for scalable text-to-image generation.
\newblock \emph{arXiv preprint arXiv:2410.08159}, 2024.

\bibitem[Gumbel(1954)]{gumbel1954statistical}
Emil~Julius Gumbel.
\newblock Statistical theory of extreme valuse and some practical applications.
\newblock \emph{Nat. Bur. Standards Appl. Math. Ser. 33}, 1954.

\bibitem[He et~al.(2016)He, Zhang, Ren, and Sun]{he2016deep}
Kaiming He, Xiangyu Zhang, Shaoqing Ren, and Jian Sun.
\newblock Deep residual learning for image recognition.
\newblock In \emph{Proceedings of the IEEE conference on computer vision and pattern recognition}, pages 770--778, 2016.

\bibitem[He et~al.(2022)He, Chen, Xie, Li, Doll{\'a}r, and Girshick]{he2022masked}
Kaiming He, Xinlei Chen, Saining Xie, Yanghao Li, Piotr Doll{\'a}r, and Ross Girshick.
\newblock Masked autoencoders are scalable vision learners.
\newblock In \emph{Proceedings of the IEEE/CVF conference on computer vision and pattern recognition}, pages 16000--16009, 2022.

\bibitem[Ho et~al.(2020)Ho, Jain, and Abbeel]{ho2020denoising}
Jonathan Ho, Ajay Jain, and Pieter Abbeel.
\newblock Denoising diffusion probabilistic models.
\newblock \emph{Advances in neural information processing systems}, 33:\penalty0 6840--6851, 2020.

\bibitem[Isola et~al.(2017)Isola, Zhu, Zhou, and Efros]{isola2017image}
Phillip Isola, Jun-Yan Zhu, Tinghui Zhou, and Alexei~A Efros.
\newblock Image-to-image translation with conditional adversarial networks.
\newblock In \emph{Proceedings of the IEEE conference on computer vision and pattern recognition}, pages 1125--1134, 2017.

\bibitem[Jain et~al.(2023)Jain, Li, Chiu, Hassani, Orlov, and Shi]{jain2023oneformer}
Jitesh Jain, Jiachen Li, Mang~Tik Chiu, Ali Hassani, Nikita Orlov, and Humphrey Shi.
\newblock Oneformer: One transformer to rule universal image segmentation.
\newblock In \emph{Proceedings of the IEEE/CVF Conference on Computer Vision and Pattern Recognition}, pages 2989--2998, 2023.

\bibitem[Kirillov et~al.(2023)Kirillov, Mintun, Ravi, Mao, Rolland, Gustafson, Xiao, Whitehead, Berg, Lo, et~al.]{kirillov2023segment}
Alexander Kirillov, Eric Mintun, Nikhila Ravi, Hanzi Mao, Chloe Rolland, Laura Gustafson, Tete Xiao, Spencer Whitehead, Alexander~C Berg, Wan-Yen Lo, et~al.
\newblock Segment anything.
\newblock In \emph{Proceedings of the IEEE/CVF International Conference on Computer Vision}, pages 4015--4026, 2023.

\bibitem[LeCun et~al.(2015)LeCun, Bengio, and Hinton]{lecun2015deep}
Yann LeCun, Yoshua Bengio, and Geoffrey Hinton.
\newblock Deep learning.
\newblock \emph{nature}, 521\penalty0 (7553):\penalty0 436--444, 2015.

\bibitem[Li et~al.(2023)Li, Chang, Mishra, Zhang, Katabi, and Krishnan]{li2023mage}
Tianhong Li, Huiwen Chang, Shlok Mishra, Han Zhang, Dina Katabi, and Dilip Krishnan.
\newblock Mage: Masked generative encoder to unify representation learning and image synthesis.
\newblock In \emph{Proceedings of the IEEE/CVF Conference on Computer Vision and Pattern Recognition}, pages 2142--2152, 2023.

\bibitem[Li et~al.(2025)Li, Tian, Li, Deng, and He]{li2025autoregressive}
Tianhong Li, Yonglong Tian, He Li, Mingyang Deng, and Kaiming He.
\newblock Autoregressive image generation without vector quantization.
\newblock \emph{Advances in Neural Information Processing Systems}, 37:\penalty0 56424--56445, 2025.

\bibitem[Liang et~al.(2023{\natexlab{a}})Liang, Wang, Zhou, Miao, Luo, and Yang]{liang2023local}
Chen Liang, Wenguan Wang, Tianfei Zhou, Jiaxu Miao, Yawei Luo, and Yi Yang.
\newblock Local-global context aware transformer for language-guided video segmentation.
\newblock \emph{IEEE Transactions on Pattern Analysis and Machine Intelligence}, 45\penalty0 (8):\penalty0 10055--10069, 2023{\natexlab{a}}.

\bibitem[Liang et~al.(2023{\natexlab{b}})Liang, Zhou, Liu, and Wang]{liang2023clustseg}
James Liang, Tianfei Zhou, Dongfang Liu, and Wenguan Wang.
\newblock Clustseg: Clustering for universal segmentation.
\newblock \emph{arXiv preprint arXiv:2305.02187}, 2023{\natexlab{b}}.

\bibitem[Nichol and Dhariwal(2021)]{nichol2021improved}
Alexander~Quinn Nichol and Prafulla Dhariwal.
\newblock Improved denoising diffusion probabilistic models.
\newblock In \emph{International conference on machine learning}, pages 8162--8171. PMLR, 2021.

\bibitem[Oquab et~al.(2023)Oquab, Darcet, Moutakanni, Vo, Szafraniec, Khalidov, Fernandez, Haziza, Massa, El-Nouby, et~al.]{oquab2023dinov2}
Maxime Oquab, Timoth{\'e}e Darcet, Th{\'e}o Moutakanni, Huy Vo, Marc Szafraniec, Vasil Khalidov, Pierre Fernandez, Daniel Haziza, Francisco Massa, Alaaeldin El-Nouby, et~al.
\newblock Dinov2: Learning robust visual features without supervision.
\newblock \emph{arXiv preprint arXiv:2304.07193}, 2023.

\bibitem[Pasini et~al.(2024)Pasini, Nistal, Lattner, and Fazekas]{pasini2024continuous}
Marco Pasini, Javier Nistal, Stefan Lattner, and George Fazekas.
\newblock Continuous autoregressive models with noise augmentation avoid error accumulation.
\newblock \emph{arXiv preprint arXiv:2411.18447}, 2024.

\bibitem[Radford et~al.(2021)Radford, Kim, Hallacy, Ramesh, Goh, Agarwal, Sastry, Askell, Mishkin, Clark, et~al.]{radford2021learning}
Alec Radford, Jong~Wook Kim, Chris Hallacy, Aditya Ramesh, Gabriel Goh, Sandhini Agarwal, Girish Sastry, Amanda Askell, Pamela Mishkin, Jack Clark, et~al.
\newblock Learning transferable visual models from natural language supervision.
\newblock In \emph{International conference on machine learning}, pages 8748--8763. PMLR, 2021.

\bibitem[Ramesh et~al.(2021)Ramesh, Pavlov, Goh, Gray, Voss, Radford, Chen, and Sutskever]{ramesh2021zero}
Aditya Ramesh, Mikhail Pavlov, Gabriel Goh, Scott Gray, Chelsea Voss, Alec Radford, Mark Chen, and Ilya Sutskever.
\newblock Zero-shot text-to-image generation.
\newblock In \emph{International conference on machine learning}, pages 8821--8831. Pmlr, 2021.

\bibitem[Razavi et~al.(2019)Razavi, Van~den Oord, and Vinyals]{razavi2019generating}
Ali Razavi, Aaron Van~den Oord, and Oriol Vinyals.
\newblock Generating diverse high-fidelity images with vq-vae-2.
\newblock \emph{Advances in neural information processing systems}, 32, 2019.

\bibitem[Rombach et~al.(2022)Rombach, Blattmann, Lorenz, Esser, and Ommer]{rombach2022high}
Robin Rombach, Andreas Blattmann, Dominik Lorenz, Patrick Esser, and Bj{\"o}rn Ommer.
\newblock High-resolution image synthesis with latent diffusion models.
\newblock In \emph{Proceedings of the IEEE/CVF conference on computer vision and pattern recognition}, pages 10684--10695, 2022.

\bibitem[Sakaridis et~al.(2021)Sakaridis, Dai, and Van~Gool]{sakaridis2021acdc}
Christos Sakaridis, Dengxin Dai, and Luc Van~Gool.
\newblock Acdc: The adverse conditions dataset with correspondences for semantic driving scene understanding.
\newblock In \emph{Proceedings of the IEEE/CVF International Conference on Computer Vision}, pages 10765--10775, 2021.

\bibitem[Shannon(1948)]{shannon1948mathematical}
Claude~E Shannon.
\newblock A mathematical theory of communication.
\newblock \emph{The Bell system technical journal}, 27\penalty0 (3):\penalty0 379--423, 1948.

\bibitem[Strudel et~al.(2021)Strudel, Garcia, Laptev, and Schmid]{strudel2021segmenter}
Robin Strudel, Ricardo Garcia, Ivan Laptev, and Cordelia Schmid.
\newblock Segmenter: Transformer for semantic segmentation.
\newblock In \emph{Proceedings of the IEEE/CVF international conference on computer vision}, pages 7262--7272, 2021.

\bibitem[Sun et~al.(2024)Sun, Jiang, Chen, Zhang, Peng, Luo, and Yuan]{sun2024autoregressive}
Peize Sun, Yi Jiang, Shoufa Chen, Shilong Zhang, Bingyue Peng, Ping Luo, and Zehuan Yuan.
\newblock Autoregressive model beats diffusion: Llama for scalable image generation.
\newblock \emph{arXiv preprint arXiv:2406.06525}, 2024.

\bibitem[Tschannen et~al.(2024)Tschannen, Eastwood, and Mentzer]{tschannen2024givt}
Michael Tschannen, Cian Eastwood, and Fabian Mentzer.
\newblock Givt: Generative infinite-vocabulary transformers.
\newblock In \emph{European Conference on Computer Vision}, pages 292--309. Springer, 2024.

\bibitem[Van Den~Oord et~al.(2017)Van Den~Oord, Vinyals, et~al.]{van2017neural}
Aaron Van Den~Oord, Oriol Vinyals, et~al.
\newblock Neural discrete representation learning.
\newblock \emph{Advances in neural information processing systems}, 30, 2017.

\bibitem[Vaswani(2017)]{vaswani2017attention}
A Vaswani.
\newblock Attention is all you need.
\newblock \emph{Advances in Neural Information Processing Systems}, 2017.

\bibitem[Wang et~al.(2023)Wang, Zhang, Cao, Wang, Shen, and Huang]{wang2023seggpt}
Xinlong Wang, Xiaosong Zhang, Yue Cao, Wen Wang, Chunhua Shen, and Tiejun Huang.
\newblock Seggpt: Towards segmenting everything in context.
\newblock In \emph{Proceedings of the IEEE/CVF International Conference on Computer Vision}, pages 1130--1140, 2023.

\bibitem[Weber et~al.(2021)Weber, Xie, Collins, Zhu, Voigtlaender, Adam, Green, Geiger, Leibe, Cremers, et~al.]{weber2021step}
Mark Weber, Jun Xie, Maxwell Collins, Yukun Zhu, Paul Voigtlaender, Hartwig Adam, Bradley Green, Andreas Geiger, Bastian Leibe, Daniel Cremers, et~al.
\newblock Step: Segmenting and tracking every pixel.
\newblock \emph{arXiv preprint arXiv:2102.11859}, 2021.

\bibitem[Weber et~al.(2024)Weber, Yu, Yu, Deng, Shen, Cremers, and Chen]{weber2024maskbit}
Mark Weber, Lijun Yu, Qihang Yu, Xueqing Deng, Xiaohui Shen, Daniel Cremers, and Liang-Chieh Chen.
\newblock Maskbit: Embedding-free image generation via bit tokens.
\newblock \emph{arXiv preprint arXiv:2409.16211}, 2024.

\bibitem[Xie et~al.(2021)Xie, Wang, Yu, Anandkumar, Alvarez, and Luo]{xie2021segformer}
Enze Xie, Wenhai Wang, Zhiding Yu, Anima Anandkumar, Jose~M Alvarez, and Ping Luo.
\newblock Segformer: Simple and efficient design for semantic segmentation with transformers.
\newblock \emph{Advances in neural information processing systems}, 34:\penalty0 12077--12090, 2021.

\bibitem[Zhang et~al.(2018)Zhang, Isola, Efros, Shechtman, and Wang]{zhang2018unreasonable}
Richard Zhang, Phillip Isola, Alexei~A Efros, Eli Shechtman, and Oliver Wang.
\newblock The unreasonable effectiveness of deep features as a perceptual metric.
\newblock In \emph{Proceedings of the IEEE conference on computer vision and pattern recognition}, pages 586--595, 2018.

\bibitem[Zheng et~al.(2021)Zheng, Lu, Zhao, Zhu, Luo, Wang, Fu, Feng, Xiang, Torr, et~al.]{zheng2021rethinking}
Sixiao Zheng, Jiachen Lu, Hengshuang Zhao, Xiatian Zhu, Zekun Luo, Yabiao Wang, Yanwei Fu, Jianfeng Feng, Tao Xiang, Philip~HS Torr, et~al.
\newblock Rethinking semantic segmentation from a sequence-to-sequence perspective with transformers.
\newblock In \emph{Proceedings of the IEEE/CVF conference on computer vision and pattern recognition}, pages 6881--6890, 2021.

\bibitem[Zou et~al.(2024)Zou, Yang, Zhang, Li, Li, Wang, Wang, Gao, and Lee]{zou2024segment}
Xueyan Zou, Jianwei Yang, Hao Zhang, Feng Li, Linjie Li, Jianfeng Wang, Lijuan Wang, Jianfeng Gao, and Yong~Jae Lee.
\newblock Segment everything everywhere all at once.
\newblock \emph{Advances in Neural Information Processing Systems}, 36, 2024.

\end{thebibliography}
}
\end{document}